\documentclass{article}

\usepackage{arxiv}
\usepackage[utf8]{inputenc} 
\usepackage[T1]{fontenc}    
\usepackage{hyperref}       
\usepackage{url}            
\usepackage{booktabs}       
\usepackage{amsfonts}       
\usepackage{nicefrac}       
\usepackage{microtype}      
\usepackage{lipsum}
\usepackage{graphicx}
\usepackage{natbib} 
\usepackage{array}
\usepackage{float}
\usepackage{caption}
\graphicspath{ {./images/} }

\title{Semantic Delta: An Interpretable Signal Differentiating Human and LLMs Dialogue}

\author{
  Riccardo Scantamburlo\\
  School of Industrial Engineering\\
  Intelligence, Complexity \\ and Technology Lab (ICT Lab)\\
  LIUC - Università Cattaneo\\
  Castellanza (Italy), 21053 \\
  \texttt{ri12.scantamburlo@stud.liuc.it} 
  \And
  Mauro Mezzanzana\\
  School of Industrial Engineering\\
  Intelligence, Complexity \\ and Technology Lab (ICT Lab)\\
  LIUC - Università Cattaneo\\
  Castellanza (Italy), 21053 \\
  \texttt{mmezzanzana@liuc.it} 
  \And
  Giacomo Buonanno\\
  School of Industrial Engineering\\
  Intelligence, Complexity \\ and Technology Lab (ICT Lab)\\
  LIUC - Università Cattaneo\\
  Castellanza (Italy), 21053 \\
  \texttt{gbuonanno@liuc.it} 
  \And
  Francesco Bertolotti \\
  School of Industrial Engineering\\
  Intelligence, Complexity \\ and Technology Lab (ICT Lab)\\
  LIUC - Università Cattaneo\\
  Castellanza (Italy), 21053 \\
  \texttt{fbertolotti@liuc.it} \\
}

\begin{document}
\maketitle
\begin{abstract}
Do LLMs talk like us? This question intrigues a multitude of scholar and it is relevant in many fields, from education to academia. This work presents an interpretable statistical feature for distinguishing human written and LLMs generated dialogue. We introduce a lightweight metric derived from semantic categories distribution. Using the Empath lexical analysis framework, each text is mapped to a set of thematic intensity scores. We define semantic delta as the difference between the two most dominant category intensities within a dialogue, hypothesizing that LLM outputs exhibit stronger thematic concentration than human discourse. To evaluate this hypothesis, conversational data were generated from multiple LLM configurations and compared against heterogeneous human corpora, including scripted dialogue, literary works, and online discussions. A Welch’s t-test was applied to the resulting distributions of semantic delta values. Results show that AI-generated texts consistently produce higher deltas than human texts, indicating a more rigid topics structure, whereas human dialogue displays a broader and more balanced semantic spread. Rather than replacing existing detection techniques, the proposed zero-shot metric provides a computationally inexpensive complementary signal that can be integrated into ensemble detection systems. These finding also contribute to the broader empirical understanding of LLM behavioural mimicry and suggest that thematic distribution constitutes a quantifiable dimension along which current models fall short of human conversational dynamics.

\keywords{LLMs, Large Language Model, AI Detection, Word Embeddings, Empath}
\end{abstract}

\section*{Introduction}
{Nowadays, Large Language Models (LLMs) are widely adopted across a variety of fields \cite{korinek2024generative}. Among their many applications, assisted writing has emerged as arguably the most prevalent use of generative AI \cite{wasi2024llms}.}

{Despite their benefits, differentiating between human and machine-generated content is essential to prevent fraudulent activity and related systemic risks \cite{king2020artificial}. Being able to tell the difference between human and AI becomes increasingly important as these models undergo iterative refinements and reinforcement learning through human feedback \cite{wu2025survey}.
This distinction also carries significant ethical implications, as it directly impacts transparency, public trust, and accountability in the use of digital technologies. Moreover, the inability to reliably identify AI-generated content may facilitate the spread of misinformation and undermine the integrity of both individual and collective decision-making processes.}

{The main AI detection techniques are: watermarking techniques, statistics-based detectors, neural-based detectors \cite{wu2025survey}.
Watermarking techniques function by embedding a hidden statistical "signature" directly into the model’s token selection process, subtly biasing the output toward a specific and identifiable distribution of vocabulary \cite{kirchenbauer2023watermark}. While this approach offers strong reliability and minimal false positives, its main weakness is its susceptibility to AI-assisted rewriting, which often obscures the subtle statistical pattern \cite{cheng2025revealing}. Furthermore, Cross-Lingual Summarization Attacks can effectively reduce detection rates to near-random levels without compromising the quality of the content \cite{ganesan2025cross}.}

{Statistics based methods detect LLM-generated text by analyzing inherent text features, without requiring supervised training or specialized access to the model. This independence makes statistics-based approaches more broadly applicable \cite{wu2025survey}. DetectGPT is one of the most well-known statistical approaches to AI-generated text detection, achieving state-of-the-art performance among methods in the same category \cite{mitchell2023detectgpt}.
Token Cohesiveness can also be used to measure the semantic closeness among tokens within a passage \cite{ma2024zero}.}

{Neural-based detectors are "black-box" tools broadly used \cite{wu2025survey}. Binary classifiers are trained on vast datasets of paired human and AI-generated samples to identify the subtle "fingerprints" of machine writing \cite{guo2023close}. While effective, these models are heavily dependent on their training data and although advances have been made, these models might still be susceptible to inherent biases \cite{tao2024cultural}. Furthermore, their high parameter count requires significant computational power \cite{strubell2019energy}. Alternatively, LLMs can function as detectors themselves; a model can analyze text to determine if it exhibits AI-generated characteristics \cite{zhu2023beat}. However, the reliability of this approach remains a subject of intense debate and is generally considered inconsistent \cite{bavaresco2025llms}.}

{Despite extensive research, current detection methods does not achieve satisfactory performance, leaving room for false positive results. In this paper, we introduce a novel statistical feature for AI detection based on semantic category distribution, utilizing the Empath library to capture thematic concentration. This approach centers on the semantic delta, defined as the numerical difference between the scores of the two most dominant thematic categories identified within a text. Our methodology takes inspiration from sentiment analysis \cite{pang2002thumbs}, \cite{baccianella2010sentiwordnet}, which showed that hidden patterns in text can be captured through lexical scoring. Rather than measuring emotional tone, however, we apply this same logic to thematic categories, tracking how topics dominate a text rather than how positive or negative they are.

Our results show that systematic analysis across a large scale sample reveals that LLM-generated texts exhibit higher deltas than human writing, suggesting that machine-generated content tends to maintain a more rigid and concentrated thematic focus. To validate these findings, a Welch’s t-test was conducted on the delta distributions, yielding a significant result; this confirms that the observed divergence is statistically significant and not the result of stochastic noise. 
Rather than serving as a standalone replacement for existing techniques, this metric is intended as a lightweight and very interpretable integration into a broader set of detection tools, which might be helpful to enhance overall accuracy. Moreover, we propose this methodology not only as a tool for detecting LLM-generated text but also as a means to discover underlying statistical patterns. This could provide a deeper understanding of how these black-box systems, whose behaviors have only recently begun to be understood \cite{gurnee2026models}, determine their responses to specific inputs."}

\begin{figure}[H]
    \centering
    \includegraphics[width=0.75\linewidth]{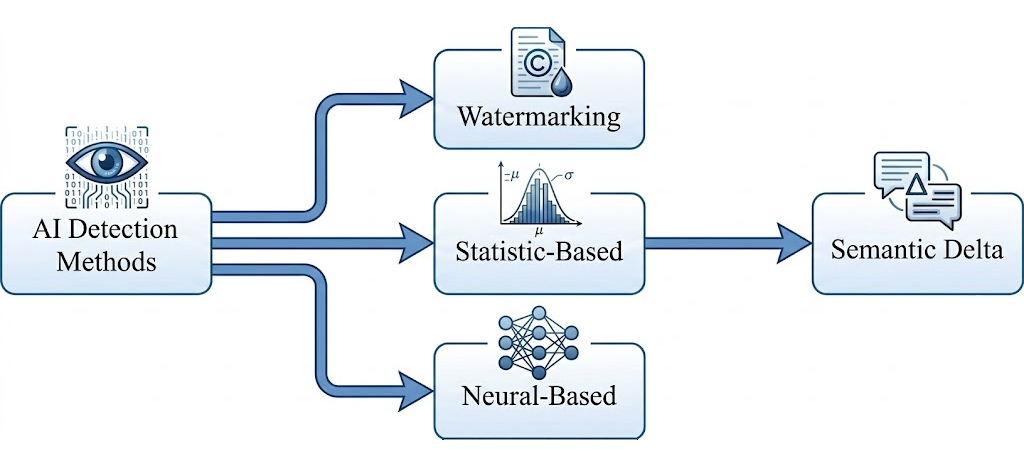}
    \caption{Our proposed metric fits into the broader AI detection methods ecosystem as a one-shot, "white-box" statistical tool.}
    \label{fig:placeholder}
\end{figure}

{The paper proceeds as it follows: first the methodology is explained, then the empirical results and the related discussion are presented. Finally, conclusions are drawn with potential future developments.}

\section*{Methods}

\subsection*{Detection process}
{Our methodology leverage on an existing open-source Python library called Empath, specifically designed to extract, through word embeddings, psychological, emotional, and topical signals from text. Empath constructs its analytical categories by projecting words into a continuous vector space. By analyzing the geometric proximity, such as cosine similarity, between these word vectors, the tool can take a small set of seed terms and automatically extrapolate closely related words to build robust semantic clusters. When processing input documents, Empath evaluates the text's vocabulary against these embedding-derived categories, returning a structured dictionary where keys are the topics and values are the raw or normalized frequency counts. This vector-based architecture allows to evaluate semantic themes within a dataset.\cite{fast2016empath}}

\begin{figure}[H]
    \centering
    \includegraphics[width=1\linewidth]{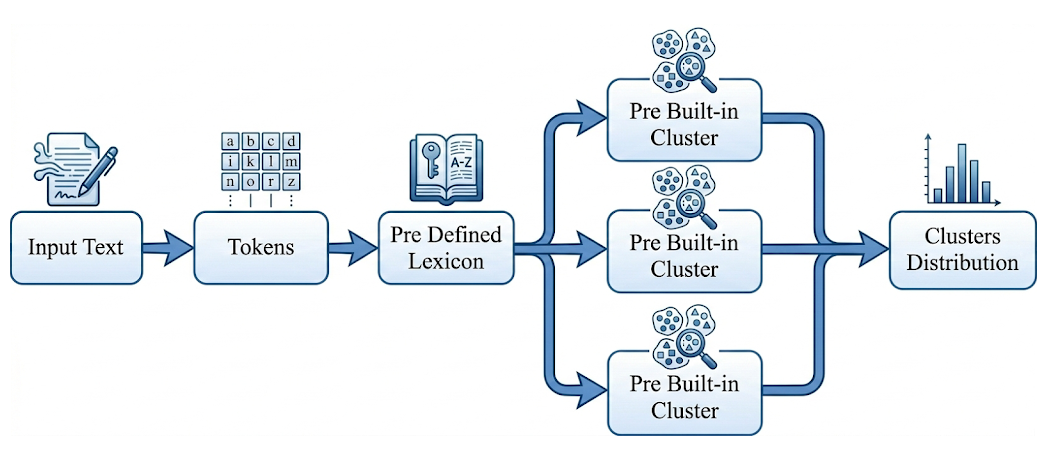}
    \caption{Lexical Analysis Pipeline using the Empath Library. After initial Tokenization, each token is matched against a Pre-Defined Lexicon consisting of 200+ Pre-Built-in Clusters. The final output represents the statistical distribution of these semantic categories within the analyzed text.}
    \label{fig:placeholder}
\end{figure}

{The proposed methodology focuses on identifying the primary categories of a text and calculating the delta, defined as the numerical difference between the value of the top-ranked category and the second-ranked category. This metric is used to test the hypothesis that AI-generated text exhibits a larger delta compared to human-generated text, suggesting that AI outputs may gravitate more toward a single dominant topic rather than a balanced distribution of themes.}

{For each given text, the main topics of conversation are identified as well as a value of intensity. This intensity value is a number, normalized with respect to the matches. 
The library divides each category count by the total number of occurrences that matched at least one lexicon category. Therefore the values express the relative share of matched indicators (not all words in the text), and the intensities may not sum to 1 because categories can overlap.}
{For each text, the two clusters with the biggest intensity value were considered. The difference $\Delta$ is computed as the difference between these two values as
$\Delta = IV_1 - IV_2$, where $IV_1$ is the intensity value of the first cluster and $IV_2$ is the intensity value of the second cluster. The primary distinction between humans and AI lies in how topics are distributed.}

\begin{figure}[H]
    \centering
    \includegraphics[width=1\linewidth]{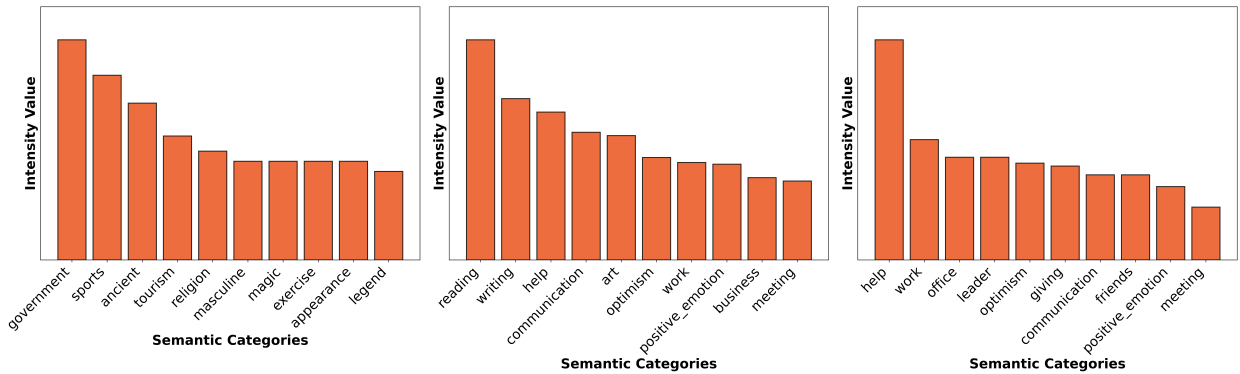}
    \caption{Empath analysis of various AI generated dialogues consistently identifies a gap of intensity value between the leading two semantic categories, suggesting a high degree of thematic concentration in the model's responses.}
    \label{fig:placeholder}
\end{figure}

\begin{figure}[H]
    \centering
    \includegraphics[width=1\linewidth]{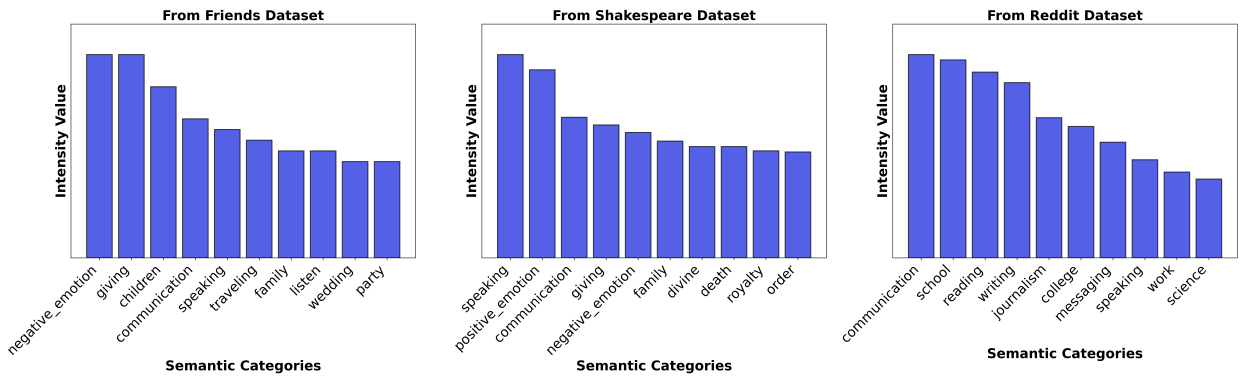}
    \caption{Empath analysis of various human-written texts consistently identifies a small gap of intensity value between the leading two semantic categories, suggesting a more dynamic and less rigid topic structure.}
    \label{fig:placeholder}
\end{figure}

In a certain way, the Semantic Delta $\Delta$ is a simpler way to measure the Shennon entropy within a text. The hypothesis is that a human conversation has higher entropy because it covers many different topics in a more balanced way. In contrast, AI writing has lower entropy because it stays focused on just one main theme \cite{scharringhausen2026entropy}. To check this, we calculated the number of bits needed to encode the information. The Shannon Entropy formula was applied to the resulting topic distribution of each text. The results showed that AI-generated text requires fewer bits than human text, proving it is more predictable and less complex, even though the difference is very small. 

\begin{figure}[H]
    \centering
    \includegraphics[width=0.75\linewidth]{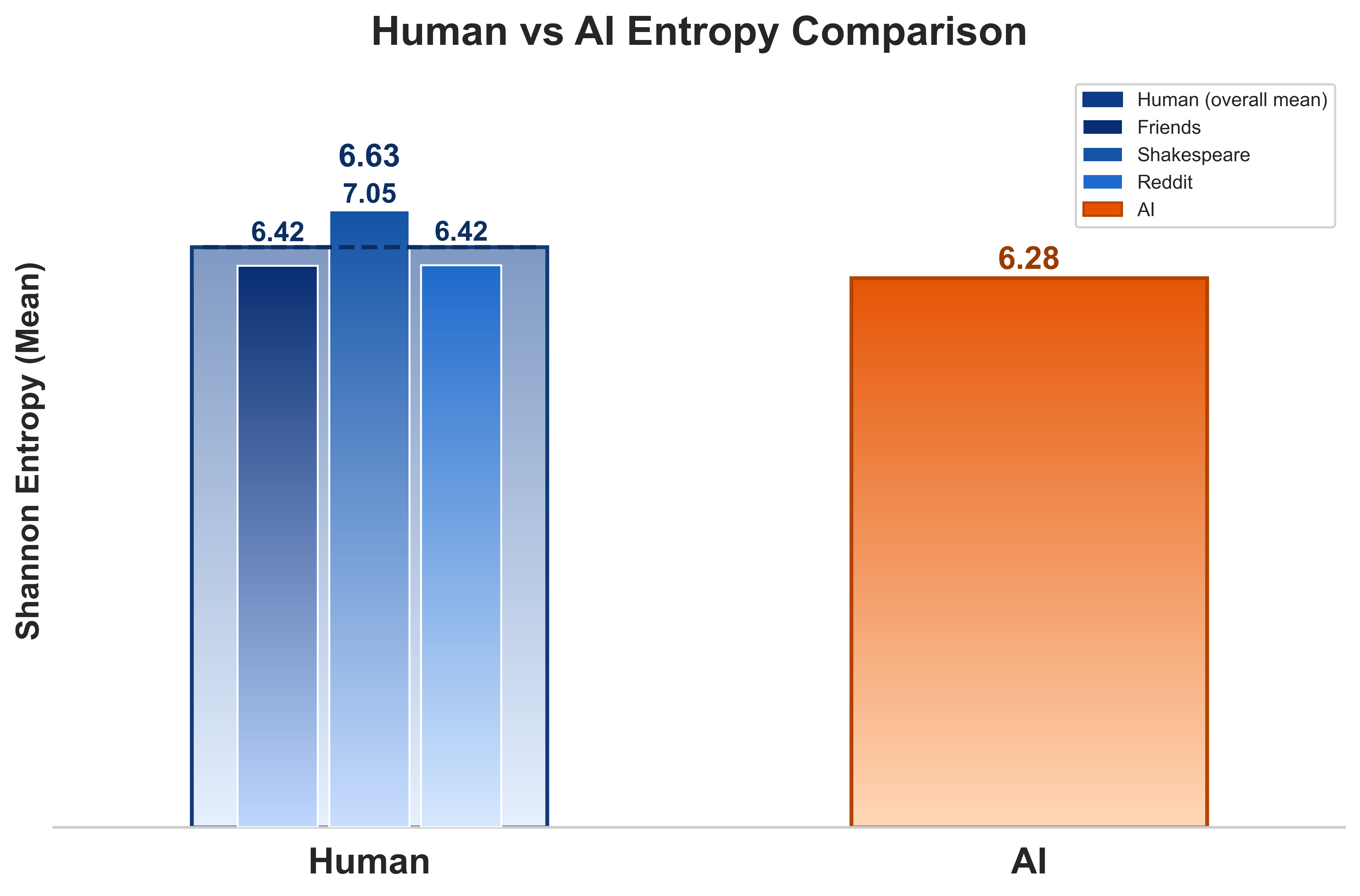}
    \caption{Comparison of mean Shannon entropy between human-written texts (Friends, Shakespeare, Reddit; aggregated as “Human”) and AI-generated texts (“AI”). The Human bar includes the three underlying human datasets to show internal variability and the overall mean. AI exhibits a slightly lower entropy than Human, indicating marginally higher predictability; although the gap is small, it is consistent with the expected trend of reduced lexical diversity in AI-generated content.}
    \label{fig:placeholder}
\end{figure}

{To assess the goodness of our intuition, a t-test was performed to test the null hypothesis that the observed difference between the mean values of the AI delta distribution and the Human deltas distribution is attributable to random noise.
Given the unequal number of AI-generated and human-generated dialogues analyzed, a Welch t-test was selected as the most appropriate approach. The sample size of AI-generated texts was sufficient to evaluate a diverse range of System and User Prompts. In contrast, a larger corpus of human dialogues was analyzed to capture the nuanced stylistic variations inherent in fiction, drama, and digital communication.}

\subsection*{Experimental settings}
{This research is built upon the analysis of interactive communication, specifically using data in the form of dialogue. Using this framework, we compare the natural spontaneity of human speech with the structured logic of Large Language Models (LLMs). Our goal is to investigate the underlying statistical patterns that reveal how these 'black-box' systems decide to respond, and to what extent they can truly mimic the complexity and naturality of human interaction."}

{The experimental framework was implemented in Python, leveraging the OpenAI API to generate LLM-based dialogues and the Empath library for thematic clustering. The library version used: 0.89. Numerical processing and data visualization were further supported by a suite of libraries: Pandas, NumPy, Math, Matplotlib, and Seaborn.}

 {The dialogue generation was executed via the OpenAI Python library. The models used are \verb|gpt-4.1-mini|, \verb|gpt-5-mini|, and \verb|gpt-4o-mini|. To capture a wide range of conversational dynamics, both system and user prompts were systematically varied across various general subjects. Initial interactions were framed using a standard "helpful assistant" system prompt. Then specialized system prompts were employed to mimic human behavior. User inputs were similarly diversified, alternating between specific directives like "Discuss about X" and open-ended requests such as "Choose a topic of discussion." This methodology ensured that the resulting dataset reflected both structured information exchange and spontaneous thematic exploration. By modulating these variables, the experiment successfully generated a diverse and suitable corpus of dialogues.}

{To represent human discourse which follows a sequential structure, three distinct datasets were sourced from Kaggle, a well-known datasets repository, each offering a unique linguistic style. The first consists of the complete scripts from the television series Friends, providing a baseline for modern, multi-party colloquial exchange. This was complemented by the full work of William Shakespeare to include a more formal, historically rich structural variety. Finally, a collection of Reddit threads regarding ChatGPT was incorporated to capture contemporary, internet-based discussions. Together, we consider these sources to provide a wide spectrum of human communication, ranging from scripted comedy and classical literature to spontaneous digital interaction, against which the AI-generated dialogues could be compared.}

{The GitHub repository associated with this study contains the source code and complete datasets necessary to replicate the experiment and check for the same results} 
\url{https://github.com/RiccardoScanta/Empath_LLM_Detection}.

\section*{Results and discussion}
{The findings of this study suggest that the semantic delta serves as an easily interpretable indicator for distinguishing between human and machine-generated discourse. The experiment demonstrated that AI models tend to exhibit a higher delta than humans. This indicates that while human communication is characterized by a more balanced and multifaceted distribution of topics, LLMs often gravitate toward a more rigid and concentrated thematic focus, probably arising as a result of models' training methods. The difference between the $Top1$ and $Top2$ topic categories is computed for each dialogue, and the resulting values are visualized as a distribution along with their mean and standard deviation.
The statistical significance of these results, validated by a Welch’s t-test with a p-value below 0.05, confirms that this divergence is not a product of stochastic noise but a reflection of inherent differences in how these entities structure conversation.
As shown in the following figure, the average for human generated texts is less than half that of AI generated texts, although the values are very small in magnitude. In terms of standard deviation, LLM generated texts exhibit a much more dispersed distribution, likely due to differences in prompting strategies, as discussed earlier.}

\begin{figure}[H]
    \centering
    \includegraphics[width=0.75\linewidth]{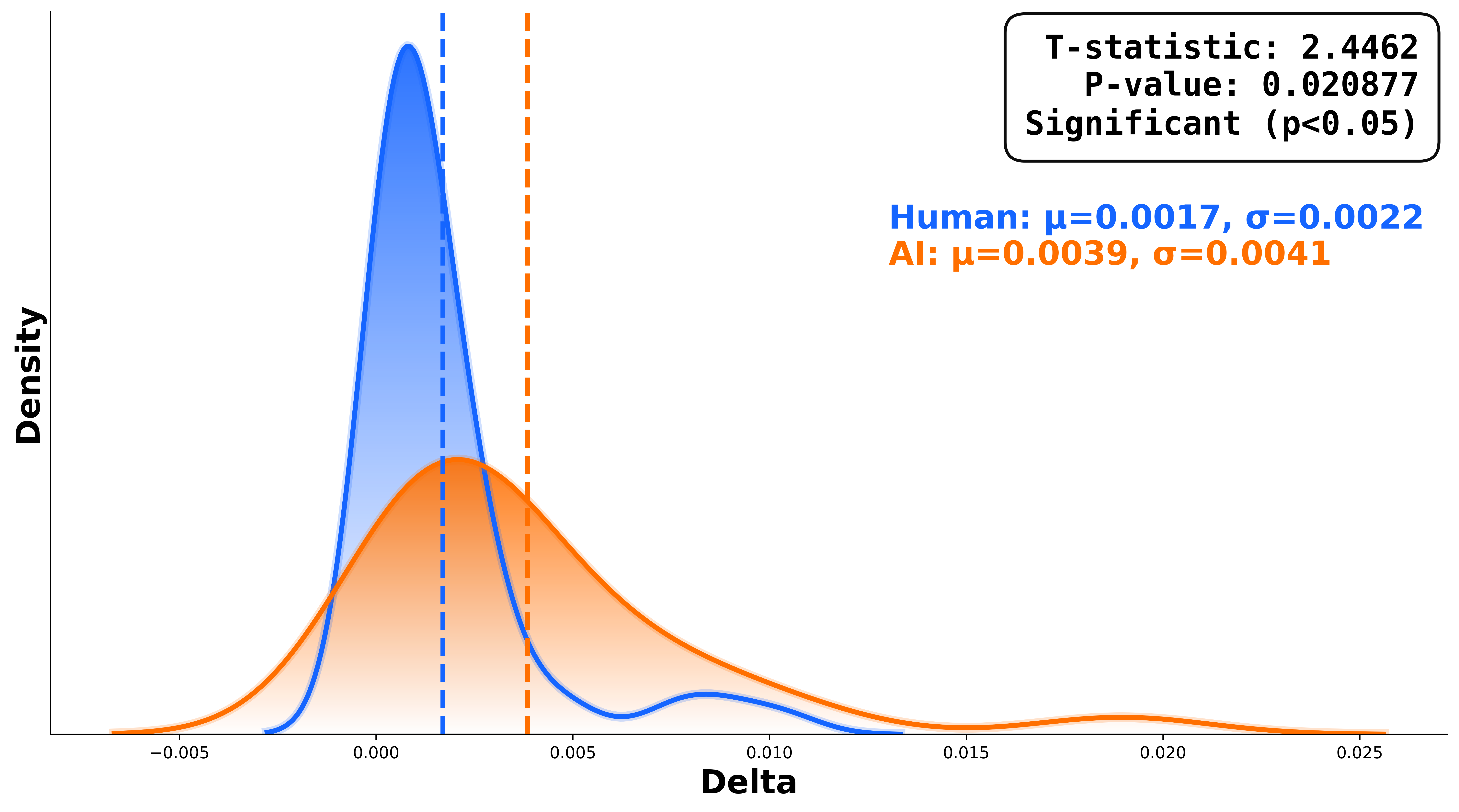}
    \caption{AI generated dialogues consistently showed a larger delta between $Top1$ and $Top2$ categories, in comparison with Human dialogues. A Welch T-Test was conducted to test the null hypothesis that the observed difference between the mean values is attributable to random noise. Since the $p <0.05$, the null hypothesis was rejected. There is a statistical significative difference between AI and Human dialogues.}
    \label{fig:placeholder}
\end{figure}

\section*{Limitations}
{Several limitations should be acknowledged. First, the human corpora used in this study, while diverse in register, are not entirely representative of spontaneous human conversation. Friends scripts and Shakespeare are authored and edited texts, and although Reddit threads capture more naturalistic dialogue, none of these sources fully replicate the unscripted dynamics of real-world social interaction. Although all of this can be considered a sufficiently comprehensive dataset, a further study could validate the findings on a larger and more various corpus. 
Second, the effect size observed, although statistically significant, is numerically small, and its practical relevance in real-world applications remains to be assessed. It has an obvious scientific interest, given that we show a statistical and simple difference between AI and human corpus, but it could be useful to a practitioner mostly as an integration. Given that it is very light computationally, this is an actual possibility. 
Third, the analysis is limited to English-language texts, and it is unclear whether the observed thematic patterns generalize across languages and cultural contexts, even if this findings are probably generalizable at least for any other languages based on a Latin alphabet.
Finally, the AI generated dialogues were produced by a specific set of OpenAI models, and the findings may not extend to other model families or architectures. This is also a limitation in term of replicability, given that analysis performed on proprietary model, at this point, can be repeated only as long as the organization that distributes them maintains them available. 
These limitations suggest caution in generalizing the results and motivate further investigation across broader and more naturalistic datasets. Addressing these open points constitutes the primary direction of future work.}

\section*{Conclusions}
{As LLMs continue to improve, the ability to distinguish AI-generated text from human writing becomes increasingly important. Understanding where and how this gap exists is not merely an academic exercise: as models become increasingly convincing, the inability to distinguish AI-generated dialogue from human conversation carries concrete risks, from the spread of misinformation to the erosion of trust in online communication. This work introduces semantic delta as a simple, interpretable signal that captures a structural difference in how humans and LLMs distribute topics across a conversation. The results show that even in open-ended dialogue settings, LLMs tend to maintain a more rigid and concentrated thematic focus than humans, and this divergence is statistically validated. This suggests that thematic distribution is a measurable dimension of conversational behaviour along which current models still differ from humans, likely as a result of their training methods. While semantic delta alone does not fully characterise human-likeness in a dialogue, it provides a lightweight and transparent framework to study it and increase our overall knowledge of the difference between human and LLM-generated text. 
Future work could focus on testing this approach across a wider range of models and larger datasets, to better understand how stable and generalizable this structural gap truly is, and whether it narrows as models continue to evolve.}

\bibliographystyle{plainnat}
\bibliography{references}  

\end{document}